\documentclass{article}

\usepackage{microtype}
\usepackage{graphicx}
\usepackage{float}
\usepackage{pifont}
\usepackage{booktabs} 
\usepackage{xspace}
\usepackage{multirow}
\usepackage{subcaption}
\usepackage[numbers]{natbib}
\newcommand{\cmark}{\textcolor{green}{\ding{51}}} 
\newcommand{\xmark}{\ding{55}} 

\usepackage{hyperref}

\usepackage[accepted]{icml2025}

\usepackage{amsmath}
\usepackage{amssymb}
\usepackage{mathtools}
\usepackage{amsthm}

\usepackage[capitalize,noabbrev]{cleveref}

\theoremstyle{plain}

\theoremstyle{definition}

\theoremstyle{remark}

\usepackage[textsize=tiny]{todonotes}

\newcommand{\frameworkname}{MAD-Sherlock}

\begin{document}
\twocolumn[
\icmltitle{\frameworkname: Multi-Agent Debate for Visual Misinformation Detection}

\icmlsetsymbol{equal}{*}

\begin{icmlauthorlist}
\icmlauthor{Kumud Lakara}{equal,ox}
\icmlauthor{Georgia Channing}{equal,ox}
\icmlauthor{Christian Rupprecht}{ox}
\icmlauthor{Juil Sock}{bbc}
\icmlauthor{Philip Torr}{ox}
\icmlauthor{John Collomosse}{surrey}
\icmlauthor{Christian Schroeder de Witt}{ox}
\end{icmlauthorlist}

\icmlaffiliation{ox}{University of Oxford, Oxford, UK}
\icmlaffiliation{bbc}{BBC AI Research, London, UK}
\icmlaffiliation{surrey}{University of Surrey, Guildford, UK}

\icmlcorrespondingauthor{Georgia Channing}{cgeorgia@robots.ox.ac.uk}
\icmlcorrespondingauthor{Christian Schroeder de Witt}{cs@robots.ox.ac.uk}

\icmlkeywords{multi-agent debate,agents,misinformation}

\vskip 0.3in
]

\printAffiliationsAndNotice{\icmlEqualContribution}


\begin{abstract}
One of the most challenging forms of misinformation involves pairing images with misleading text to create false narratives. Existing AI-driven detection systems often require domain-specific finetuning, limiting generalizability, and offer little insight into their decisions, hindering trust and adoption. We introduce \frameworkname{}, a multi-agent debate system for out-of-context misinformation detection.
\frameworkname{} frames detection as a multi-agent debate, reflecting the diverse and conflicting discourse found online. Multimodal agents collaborate to assess contextual consistency and retrieve external information to support cross-context reasoning.
Our framework is domain- and time-agnostic—requiring no finetuning\textendash yet achieves state-of-the-art accuracy with in-depth explanations. Evaluated on NewsCLIPpings, VERITE, and MMFakeBench, it outperforms prior methods by 2\%, 3\%, and 5\%, respectively.
Ablation and user studies show that the debate and resultant explanations significantly improve detection performance and improves trust for both experts and non-experts, positioning \frameworkname{} as a robust tool for autonomous citizen intelligence.
\end{abstract}

\section{Introduction}
\label{introduction}
The rise of online news and social media has been paralleled by a surge in digital misinformation~\citep{intro_cite_1,intro_cite_2,intro_cite_3}. Among the most widespread tactics is out-of-context (OOC) image use~\citep{pbsOutofcontextPhotos}, where unaltered images are paired with misleading text to deceive, requiring little technical skill.

OOC detection demands nuanced reasoning to identify misalignment between images and accompanying text. This task is time-consuming for humans, and detection accuracy drops under time pressure~\citep{sultan2022time}, limiting scalability.

AI tools offer a path forward, but traditional forensic methods~\citep{castillo2021comprehensive, heidari2024deepfake, zhu2018deep, amerini2021deep, hina2021sefaced} focus on tampering artifacts (e.g., Photoshop edits~\citep{photo_forensics, Wang_2019_ICCV} or Deepfakes~\citep{tolosana2020deepfakes, masood2023deepfakes, mitDeepfakesExplained}) and are ill-suited for OOC detection, which hinges on cross-contextual reasoning rather than visual manipulation.

\begin{figure*}[h]
    \begin{center}
    \includegraphics[width=0.72\linewidth]{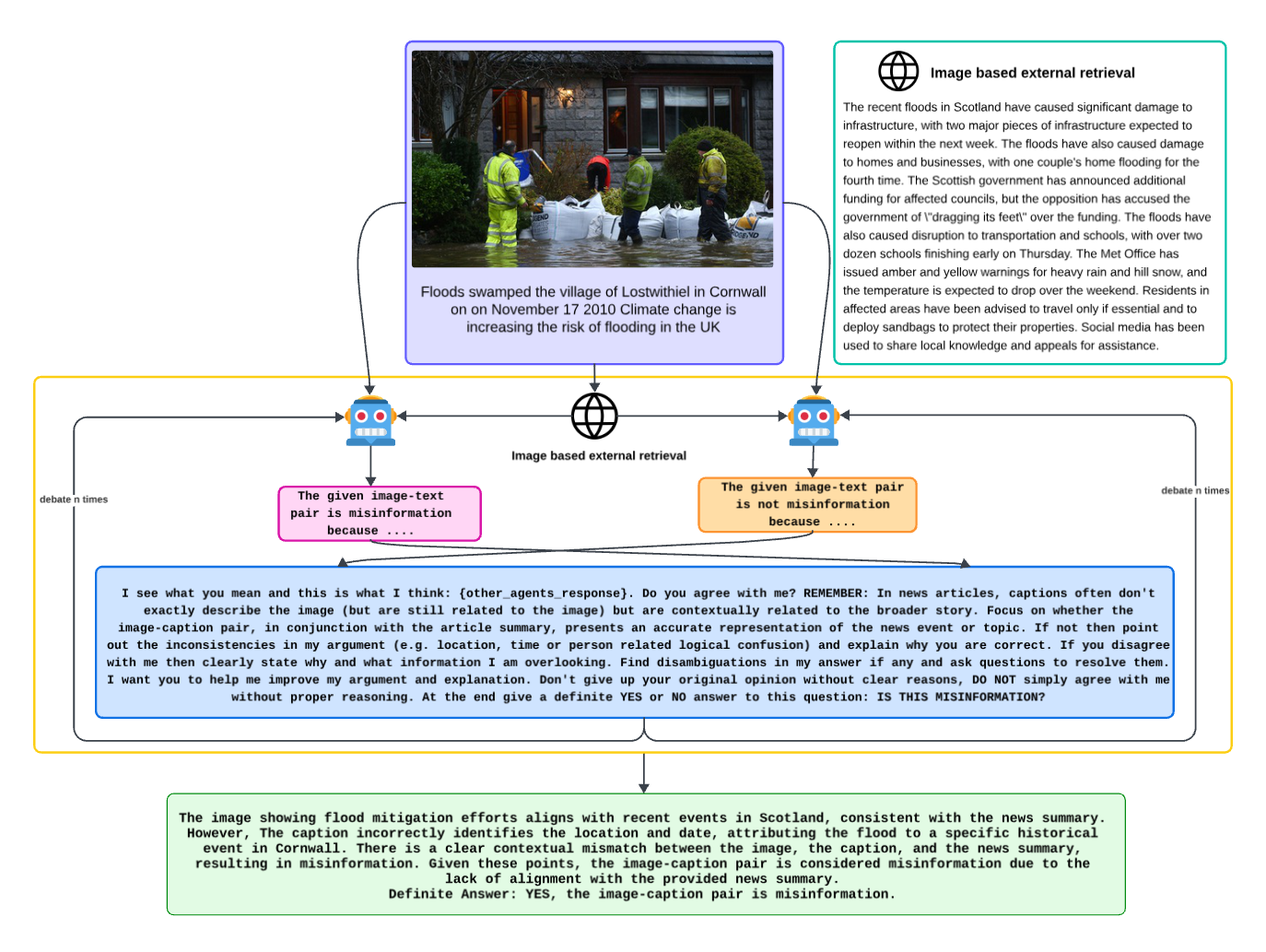}
    \end{center}
    \caption{\textbf{Overview of \frameworkname{}:} Two or more independent agents see the same image-text input and are tasked with detecting whether the input is misinformation or not. After the agents form their independent opinions, they participate in a debate until they converge on the same response or when $n$ debate rounds are completed (whichever is earlier).}
    \label{fig:main_figure}
\end{figure*}

Pretrained large multimodal models (LMMs) offer a promising foundation for detecting OOC image use by jointly interpreting text and images~\citep{liu2024llavanext, openai2024gpt4technicalreport, li2019visualbert, radford2021learning}. However, applying them to news content is challenging. News articles often feature loosely related but contextually appropriate images—e.g., a pre-2024 photo of Donald Trump used in election coverage—which makes OOC detection difficult using pretraining alone.
LMMs also suffer from hallucinations, misinterpret intent~\citep{bai2024hallucinationmultimodallargelanguage, liu2024mitigatinghallucinationlargemultimodal}, and lack up-to-date knowledge. Off-the-shelf models exhibit these limitations, reducing their reliability. Fine-tuning helps~\citep{qi2024sniffermultimodallargelanguage}, but is costly and requires frequent updates. Crucially, beyond detection, models must provide clear, human-readable explanations to support understanding and trust in their decisions.

In this work, we propose a novel post-training framework for scalable OOC misinformation detection that improves contextual reasoning, offers built-in explainability, and achieves state-of-the-art accuracy without task-specific fine-tuning (see Section~\ref{sec:method}). Our method frames the detection problem as a dialectic debate between LMM agents, augmented with external information retrieval.

Unlike single-agent chain-of-thought methods~\citep{wei_chain--thought_2024}, our multi-agent setup enables context separation, decentralized reasoning, and parallelization~\citep{de_witt_is_2020,du_improving_2023}. Prior work~\citep{can_llms_keep_a_secret} shows agents struggle to maintain diverse perspectives within a single context window; our approach addresses this via modular, compositional interaction. It also allows seamless integration of human or autonomous agents, making \frameworkname{} a flexible tool for expert oversight. To our knowledge, this is the first use of multi-agent LMM debate for both detecting and \textit{explaining} OOC image use.

\frameworkname{} is backbone-agnostic, compatible with open- and closed-source models. We prototype and tune with LLaVA~\citep{liu2024llavanext} to reduce API costs, then deploy GPT-4o~\citep{gpt4o} for final results. By avoiding task-specific fine-tuning, our method remains broadly applicable across domains and time periods.

We evaluate on three benchmarks—NewsCLIPpings, VERITE, and MMFakeBench—and achieve new state-of-the-art performance across all. \frameworkname{} outperforms prior methods and baselines, demonstrates robustness, and produces detailed, interpretable explanations. In user studies, these explanations significantly improve detection accuracy for both experts and non-experts.

Key contributors to performance include external retrieval and agent independence. We discuss limitations and outline future directions toward scalable, general-purpose AI for public good.

\section{Related Work}

Recent work has explored joint image-text representations for OOC classification. \citet{aneja2022acmmultimediagrandchallenge} use a self-supervised approach to enforce image-text alignment via an object-caption matching score, classifying OOC instances by caption similarity. However, the method is heavily text-dependent and lacks interpretability.

\citet{abdelnabi2022opendomaincontentbasedmultimodalfactchecking} introduce the Consistency Checking Network (CCN), which leverages memory networks and CLIP for image-caption consistency against external evidence, improving accuracy but offering only binary outputs without explanations.

\citet{zhang2024interpretabledetectionoutofcontextmisinformation} employ AMR-based symbolic graphs for interpretable OOC detection with evidence. Similarly, \citet{zhou2020safesimilarityawaremultimodalfake} propose SAFE for joint text-visual reasoning in fake news detection. \citet{eann} present EANN, which uses adversarial training to extract event-invariant multimodal features. These methods require training from scratch and lack the advanced reasoning and knowledge of large pretrained models.


Sniffer~\citep{qi2024sniffermultimodallargelanguage} is most similar to our work. It uses InstructBLIP~\citep{dai2023instructblipgeneralpurposevisionlanguagemodels} to detect OOC image use and generate explanations by aggregating internal and external knowledge from entity extraction APIs and image-based web searches. However, its external knowledge is limited to basic textual content (e.g., article titles), and adapting it to the news domain requires extensive training, reducing generalizability and increasing computational overhead.

The CRAVE framework~\citep{crave} clusters retrieved multimodal evidence into coherent narratives, then uses an LLM to produce fact-checking judgments with natural language explanations. While effective, treating all clusters equally may amplify fringe perspectives and distort the evidence base.

\citet{mmfakebench} introduce MMD-Agent, a single-agent framework that hierarchically decomposes misinformation detection into textual, visual, and cross-modal subtasks before reasoning over their outputs to make a final prediction.

\section{Methodology}
\label{sec:method}
We present \frameworkname{}, an explainable misinformation detection system that jointly predicts and explains instances of misinformation (Figure \ref{fig:main_figure}). Unlike prior work, which largely provides predictions without explanations, our approach uses multiple multi-modal models debating to determine whether an image-text pair constitutes misinformation. We address the question:  

\begin{center}  
\textit{Can debating multi-modal models, equipped with external context, detect misinformation by identifying subtle contextual inconsistencies?}  
\end{center}  

Our external retrieval module uses reverse image search to provide agents with real-world context, enhancing their predictions. Using the GPT-4o \cite{gpt4o} model, we achieve state-of-the-art performance with detailed, coherent explanations, without requiring domain-specific fine-tuning. This ensures faster generalization to new domains with minimal computational overhead.

\subsection{Debate Modelling}
Analogous to real-world conversations, communication between two AI agents can also be structured in a myriad of ways. We explore multiple debating strategies to structure the conversation between agents, all of which are tested and evaluated in our experiments and informed by the work of~\citet{browncohen2023scalableaisafetydoublyefficient}. 
\begin{figure*}[t]
    \centering
    \includegraphics[width=0.8\linewidth]{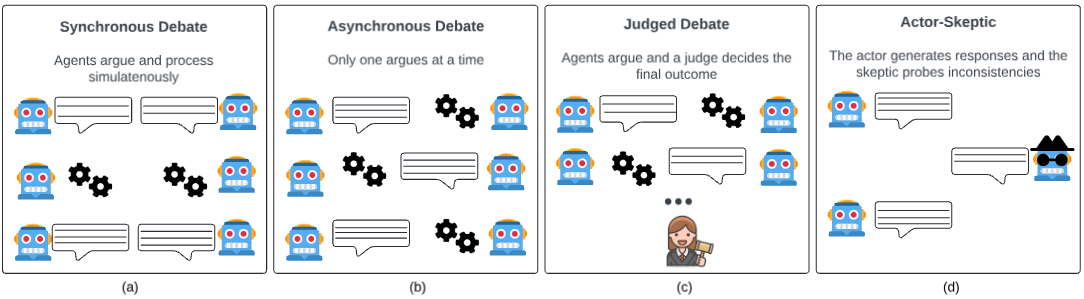}
    \caption{\textbf{Debating Strategies:} We experiment with multiple debating strategies to evaluate which performs best on our task.}
    \label{fig:debate_strategies}
\end{figure*}
 
\paragraph{Asynchronous Debate (not) Against Human:} We experiment with an asynchronous debating strategy, where models wait for others' responses before generating their own (Figure~\ref{fig:debate_strategies}a vs.~b). While synchronous debates are faster and more efficient, the asynchronous setup proves more effective for identifying contextual ambiguities—critical in misinformation detection. Notably, models are prompted to believe they are debating a human rather than another AI.

\paragraph{Judged Debate:}We also explore an asynchronous debate setup with a judge (Figure~\ref{fig:debate_strategies}c), where models debate as usual, but a final decision is made by a judge based solely on the debate transcript. Following~\citet{khan2024debatingpersuasivellmsleads}, the judge lacks access to external information, incentivizing models to structure arguments that are maximally persuasive.

\paragraph{Actor-Skeptic:}In this setup, a single \textit{actor} determines whether an image-text pair constitutes misinformation. A \textit{skeptic} then critiques the actor’s reasoning, probing for logical flaws and ambiguities. Neither agent has access to the ground truth, and since their roles are distinct, this configuration does not benefit from ensembling.

\subsection{Prompt Engineering}\label{sec:prompt_engg}

The debate is structured via prompt engineering (Figure~\ref{fig:main_figure}). In the first stage, each agent independently assesses whether the image-text pair is misinformation, using context from an external retrieval module (see Appendix~\ref{app:prompts}). Prompts summarize relevant articles and emphasize visual cues (e.g., watermarks, flags) to guide initial judgments. Agents then debate: the first round uses a tailored prompt to initiate discussion; later rounds use a standard prompt incorporating prior responses. Agents must agree or disagree with peers while refining reasoning, and prompts explicitly discourage blind agreement by requiring justification for any alignment.

\subsection{External Information Retrieval}
\begin{figure*}[h]
    \centering
    \includegraphics[width=0.65\linewidth]{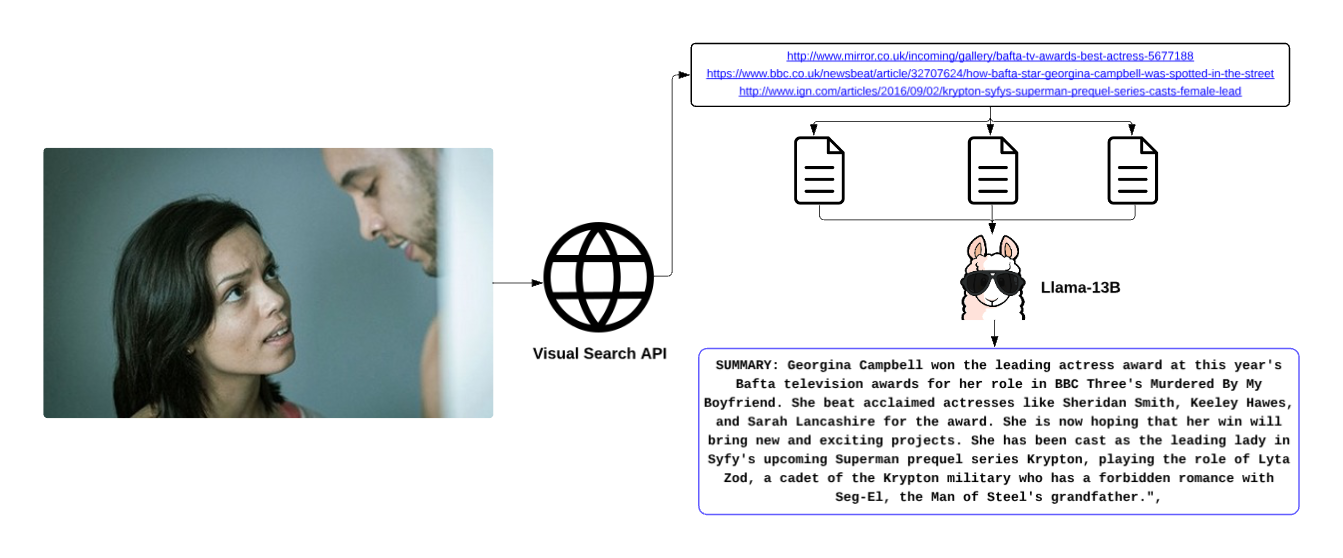}
    \caption{\textbf{Structure of the external information retrieval module}: We use the Bing Visual Search API to obtain web pages related to a given image, which are then summarised using Llama-13B \citep{llama}. This summary is then passed to the debating agents as a part of the initial prompt.}
    \label{fig:enter-label}
\end{figure*}

A model’s world knowledge is bounded by its training data and time frame, but external retrieval enables access to up-to-date context. Prior work leverages retrieval-based datasets \citep{abdelnabi2022opendomaincontentbasedmultimodalfactchecking}, though these typically include only news headlines, which are often too sparse for reliable inference. Full article access can significantly improve misinformation detection. To this end, we introduce a two-stage external retrieval module that enhances accuracy when integrated into the pipeline:

\paragraph{API-Based Information Retrieval}
We use the Bing Visual Search API to retrieve web pages related to each image. For each image, we collect the top three matching pages in which it appears, assuming these provide sufficient context for understanding the image’s original use. Since NewsCLIPpings \citep{luo2021newsclippings} includes articles over a decade old, some images yield no search results. In these rare cases, we omit external context and rely solely on the agent’s prior knowledge, which has minimal impact on overall performance.

\paragraph{Summarization using LLM}
After retrieving the top three web pages, we scrape their text and use LLaMA-13B \citep{llama} to generate concise summaries focused on key contextual details. Since the model struggles with non-English content, we filter out non-English pages--limiting the system to English but without affecting performance, as the dataset is predominantly English. Multilingual support could be added via translation prior to summarization.

\paragraph{Coherent Reasoning}
This stage of the pipeline integrates all components of \frameworkname{}. Each multimodal agent engages in the optimal debate setup using relevant prompts and is tasked with determining and explaining whether a given image-text pair constitutes misinformation. Agents also access contextual information via the external retrieval module. The system’s final decision is produced when the debate concludes, which is either after a fixed number of rounds or once all agents converge on a shared response, whichever occurs first.

\section{Experiments and Results}
\label{sec:exp}
\subsection{Dataset}
We conduct experiments on three datasets: NewsCLIPpings, VERITE, and MMFakeBench~\citep{luo2021newsclippings, verite, mmfakebench}. This selection balances well-established benchmarks like NewsCLIPpings with newer, more diverse datasets such as VERITE and MMFakeBench, providing both continuity with prior work and coverage of recent advances.

The NewsCLIPpings dataset builds on VisualNews~\citep{liu2020visualnews}, which contains image-caption pairs from BBC, USA Today, The Guardian, and The Washington Post. OOC samples are created by replacing an image with a semantically related one from a different pair. We use the Merged-Balanced version, with 71,072 training, 7,024 validation, and 7,264 test samples, offering balanced retrieval strategies and label distributions. Its scale and prior adoption make it well-suited for evaluating our model’s ability to detect out-of-context content.

We also evaluate on the VERITE dataset~\citep{verite}, a recent benchmark for multimodal misinformation detection that controls for unimodal bias. Each image-caption pair is constructed so that neither modality alone determines veracity. VERITE includes three subsets—All, True vs.~OOC, and True vs.~MC—each testing different aspects of multimodal reasoning. Captions and images appear in both truthful and misleading contexts, promoting balanced, joint reasoning. The test split contains 338 true, 324 OOC, and 338 MC samples. We report results on the combined misinformation set (MC + OOC), with disaggregated results in Appendix~\ref{app:results}.

We also evaluate on MMFakeBench~\citep{mmfakebench}, a benchmark for multimodal misinformation detection in mixed-source scenarios. The validation set includes 1,000 image-caption pairs spanning three major categories—textual, visual, and cross-modal distortions—each with 12 subtypes, capturing the complexity of real-world misinformation.

\subsection{Experimental Setup}
All experiments were run on 8 A40 (46GB) Nvidia GPU server. The estimated cost of processing one data sample using \frameworkname{} with a GPT-4o backbone is \$0.24. Inference times range from 5 to 15 seconds.
\paragraph{Debate Setup} We conduct experiments to select the best debating configuration using the LLaVA model \citep{liu2024llavanext}. All experiments are run for $k=3$ rounds or until the agents converge (whichever is earlier).
\paragraph{External Retrieval Module} We use the Bing Visual Search API to run an image-based reverse search. Using the API we select the top $k=3$ pages in which the image appears and scrape the text from them using the \texttt{Newspaper3k} library. Finally, we use Llama-13B \citep{llama} to summarise the text obtained from the top $k=3$ web pages. This step is crucial since the web pages are usually news articles which contain large amounts of text which, when scraped and passed directly to the model, can exceed its maximum token length.
\paragraph{Baselines and Prior Work}\label{baselines} The models are presented with the image and caption pair and asked if the pair is misinformation. The models are further prompted to explain their reasoning. We present comparisons to other explainable methods, specifically Sniffer~\citep{qi2024sniffermultimodallargelanguage}, CRAVE~\citep{crave}, and MMD-Agent~\cite{mmfakebench} in the following section. We also compare \frameworkname{} to existing pretrained multi-modal baselines including CLIP~\citep{radford2021learning}, VisualBERT~\citep{li2019visualbert}, InstructBLIP~\citep{dai2023instructblipgeneralpurposevisionlanguagemodels}, LLaVA~\citep{liu2024llavanext}, and GPT-4o~\citep{openai2024gpt4technicalreport, gpt4o} on the NewsCLIPpings dataset in Appendix~\ref{app:results}. We show results for two baseline methods trained from scratch on the NewsCLIPpings, namely EANN~\citep{eann} and SAFE~\citep{zhou2020safesimilarityawaremultimodalfake}, in Appendix~\ref{app:results}. We further compare \frameworkname{} to DT-Transformer \citep{papdopoulos}, CCN \citep{abdelnabi2022opendomaincontentbasedmultimodalfactchecking}, VINVL \citep{huang2024exposing}, SSDL \citep{Mu_2023_WACV}, and Neuro-Sym \citep{zhu2022generalizationdifferencesendtoendneurosymbolic} in Appendix~\ref{app:results}. 

\subsection{Results}
We present results for the experiments conducted to select the best debate setup as well compare the performance of \frameworkname{} against existing methods. We use classification accuracy as the primary performance metric for comparison based on quantitative analysis.
\subsubsection{Comparing Debate Setups}
We compare multiple debating setups using the LLaVA-v1.6-34B model, to select the best one for comparison with other works and further experimentation. 
\begin{table}[h]
\centering
\resizebox{0.99\linewidth}{!}{%
\begin{tabular}{l c c c}
\toprule
\multicolumn{1}{l}{\textbf{Debate Setup}} & \multicolumn{1}{c}{\textbf{Accuracy}} & \multicolumn{1}{c}{\textbf{Precision}} & \multicolumn{1}{c}{\textbf{Recall}} \\
\midrule
Actor-Skeptic & 69.5 & 66.1 & 69.4 \\
Judged Debate & 66.7 & 66.7 & 61.5 \\
\texttt{Async\_Debate\textsubscript{AI}} (believes debating AI) & 75.2 & 54.5 & 86.4 \\
\texttt{Async\_Debate\textsubscript{human}} (w/o external info) & 77.1 & 68.4 & 89.3 \\
\texttt{Async\_Debate\textsubscript{human}}(w external info) & \textbf{86.2} & \textbf{82.6} & \textbf{90.6} \\
\bottomrule
\end{tabular}
}
\vspace{0.2cm}
\caption{\textbf{Performance comparison between different debate setups: } We observe best performance with the \texttt{The Async\_Debate\textsubscript{human}} set-up, in which the model has external context and believes it is debating a human.}
\label{tab:setup_comparison}
\end{table}

Table~\ref{tab:setup_comparison} shows that \texttt{Async\_Debate\textsubscript{human}} with external information outperforms all other debate configurations. To highlight the role of external retrieval, we also report results without it and observe a significant performance drop.

We find that agents perform better when they believe they are debating a human, engaging more critically with peer responses. The asynchronous setup also benefits from ensemble effects, unlike the actor-skeptic setup where only one agent generates responses.

In judged debates, agents must adhere to fixed stances and persuade a judge, which can lead to confusion and errors. In contrast, \texttt{Async\_Debate\textsubscript{human}} allows agents to revise their views mid-debate, leading to clearer outcomes. Based on these findings, we adopt this setup—with external information—for all further experiments.

\subsubsection{Performance Comparison}
We present our results on the NewsCLIPpings, VERITE, and MMFakeBench datasets against existing out-of-context detection methods discussed in Section~\ref{baselines}. We include additional comparisons against legacy methods in Appendix~\ref{app:results}.

\begin{table}[t]
    \centering
    \resizebox{0.99\linewidth}{!}{%
    \begin{tabular}{l l c c c}
        \toprule
        \multirow{2}{*}{\textbf{Dataset}} & \multirow{2}{*}{\textbf{Model}} & \multicolumn{3}{c}{\textbf{Performance}} \\
        \cmidrule(lr){3-5}
        & & \textbf{Accuracy} & \textbf{Precision} & \textbf{Recall} \\
        \midrule
        \multirow{3}{*}{\textbf{NewsCLIPpings}} 
        & SNIFFER~\cite{qi2024sniffermultimodallargelanguage} & 88.4 & 91.8 & 86.9  \\  
        & CRAVE~\citep{crave} & 85.0 & 85.0 & 85.0 \\
        & \frameworkname{} (ours) & \textbf{90.8} & 85.5 & 99.0 \\
        \midrule
        \multirow{3}{*}{\textbf{VERITE}} 
        & CRAVE~\citep{crave} & 82.0 & 82.0 & 82.0 \\
        & \frameworkname{} (ours) & \textbf{85.2} & 83.6 & 96.0 \\
        \midrule
        \multirow{3}{*}{\textbf{MMFakeBench}} 
        & MMD-Agent~\cite{mmfakebench} & 62.1 & 67.8 & 59.3 \\
        & CRAVE~\citep{crave} & 78.0 & 83.6 & 69.7 \\
        & \frameworkname{} (ours) & \textbf{83.8} & 87.2 & 90.1 \\
        \bottomrule
    \end{tabular}}
    \vspace{0.1cm}
    \caption{\textbf{Performance comparison of LLM-Consensus against other explainable methods on various datasets:} Performance of LLM-Consensus and baselines across NewsCLIPpings, VERITE, and MMFakeBench. Our model shows consistent generalization and superior performance across datasets.}
    \label{tab:cross-dataset-results}
\end{table}

Table \ref{tab:cross-dataset-results} shows the comparison between our method and other explainable methods on the NewsCLIPpings, VERITE, and MMFakeBench datasets. We report state-of-the-art performance when using our proposed debate configuration with the GPT-4o \citep{openai2024gpt4technicalreport, gpt4o} model across all datasets. As a reminder, Sniffer \citep{qi2024sniffermultimodallargelanguage} is finetuned extensively to adapt it to the NewsCLIPpings dataset. MMD-Agent represents the single-agent framework proposed by the authors of the MMFakeBench dataset~\cite{mmfakebench}. The CRAVE framework introduced by~\citet{crave} represents the most recent work across all three datasets, but our framework shows markedly superior performance across all datasets.

We find that our system produces coherent, detailed and comprehensive explanations. We also note that the debate paradigm in itself is essential to the system performance. We observe a drop in performance and quality of explanations when using an identical system configuration but with a single model. We believe that this follows from work by~\citet{can_llms_keep_a_secret}, which introduces and demonstrates the importance of contextual privacy. Introducing multiple context windows allows each agent to maintain its own role and informational perspective without leakage, which is not possible in single-agent systems.

We provide a qualitative sample of our framework's explanations in Appendix~\ref{app:qualitative}. We provide ablation studies on each component of the \frameworkname{} pipeline in Appendix~\ref{app:ablations}.

We also note that single multi-modal models that do not do retrieval, including VisualBERT, CLIP, InstructBLIP, LLaVA and GPT-4o do not perform at par with other related work. We demonstrate these results in Appendix~\ref{app:results}. This can be attributed to the necessity for external context for misinformation detection and the lack of diverse perspectives that occur naturally in a multi-agent framework. These models require additional integration into more comprehensive pipelines, as done in this work.


\section{User Study}
We conducted a user study to assess our system’s ability to detect and explain misinformation—particularly important given the lack of standard metrics for evaluating explanation quality. Participants were grouped by profession: Journalists, Academics, and Others (see Appendix~\ref{app:user_study}).

Each participant reviewed ten image-text pairs, judged their veracity, and rated their confidence (0–10). After submitting initial responses, they viewed \frameworkname{}’s explanations and revised their answers. As shown in Table~\ref{tab:user_study_table}, \frameworkname{} outperformed average human accuracy both with and without AI assistance, highlighting its potential for improving public safety and trust.

\begin{table}[t]
\centering
\label{user_study_table}
\resizebox{0.8\linewidth}{!}{
\begin{tabular}{lc}
\toprule
\bf Study Setup &\multicolumn{1}{c}{\bf Average Accuracy} \\
\midrule
Humans        &$60.3\pm13.5$ \\
Humans+\frameworkname{}     & $76.7\pm12.2$ \\
\textbf{\frameworkname{}}   
&$\mathbf{80.0\pm0.0}$ \\
\bottomrule
\end{tabular}}
\vspace{0.1cm}
\caption{\textbf{Performance comparison between different study setups: }\frameworkname{} outperforms humans with and without AI assistance.}
\vspace{-0.25cm}
\label{tab:user_study_table}
\end{table}

Group-wise analysis, shown in Table \ref{tab:group_comparison}, reveals significant performance gains across all groups, with results approaching those of professional journalists. Confidence levels (out of 10) are comparable across groups and generally increase after using \frameworkname{} insights. Thus, \frameworkname{} can substantially boost non-expert performance, making it valuable for citizen intelligence applications.

\begin{table}[t]
\centering
\setlength{\tabcolsep}{3pt}
\renewcommand{\arraystretch}{1.2}

\resizebox{0.99\linewidth}{!}{
\begin{tabular}{lccc}
\toprule
\textbf{Metric} & \textbf{Journalists} & \textbf{Academics} & \textbf{Others} \\ 
\midrule
Accuracy (only human) & $70.0\pm1.4$ & $60.7\pm1.4$ & $56.7\pm1.5$ \\
Confidence (only human) & $4.3\pm2.1$ & $3.2\pm0.8$ & $3.9\pm1.2$ \\
Accuracy (with \frameworkname{}) & $82.2\pm0.9$ & $79.3\pm1.3$ & $71.7\pm1.1$ \\
Confidence (with \frameworkname{}) & $5.3\pm1.3$ & $5.8\pm1.4$ & $5.8\pm1.4$ \\
\bottomrule
\end{tabular}}

\vspace{0.2cm}
\caption{\textbf{Performance comparison:} \frameworkname{} improves performance across all participant groups.}
\label{tab:group_comparison}
\end{table}

\section{Conclusion and Future Work}

Out-of-context (OOC) image misuse is an increasing challenge for misinformation detection, especially as vision-language models grow more powerful yet less interpretable. We explore whether multiple AI agents can collaboratively reason about context to improve prediction accuracy. Our strongest results come from the \texttt{Asynchronous\_Debate\textsubscript{human}} setup, where agents believe they are debating a human. This setting promotes engagement, mid-debate revision, and better identification of subtle inconsistencies.

Our final system, \frameworkname{}, achieves state-of-the-art performance while offering interpretable, evidence-based explanations—enabled by our external retrieval module. We observe substantial gains in OOC detection across both expert and non-expert users.

We see several avenues for future work, including a continuously updated benchmark with recent news and more nuanced inconsistencies, extending to video-text and multilingual inputs, and large-scale deployments in professional and citizen intelligence settings. We hope to engage with the community to better understand how agentic workflows can enhance online safety. For limitations, see Appendix~\ref{app:limitations}.

\pagebreak
\bibliographystyle{plainnat}  
\bibliography{icml}

\newpage
\appendix
\onecolumn

\newpage
\section{Appendix}\

\subsection{Limitations}
\label{app:limitations}

Despite the strong performance of \frameworkname{}, several limitations remain. First, while our model excels at detecting out-of-context image–text pairs, its reliance on external retrieval can lead to reduced accuracy when relevant context is unavailable or difficult to retrieve. Moreover, our framework cannot independently verify the factual correctness of externally retrieved news articles; the truthfulness of any source may be debated, potentially introducing misinformation into the reasoning process. Nevertheless, we observe that the model’s judgments align closely with the human-created labels used in these widely accepted datasets, underscoring its practical utility. Second, the quality of explanations is constrained to textual outputs, limiting multi-modal explanation capabilities such as image or video integration. Third, the system’s performance is sensitive to hyperparameter tuning, including the number of debate rounds and agents, which may require further optimization for broader use cases.

Additionally, while our user studies provided valuable insights, large-scale deployment in diverse, real-world settings, such as professional or citizen intelligence environments, is necessary to fully assess the method’s robustness and scalability. Finally, our dataset, though comprehensive, primarily focuses on English-language news, limiting the generalizability of the system across non-English contexts.

Another important limitation is the potential risk that open-sourcing \frameworkname{} might allow adversaries to train models specifically designed to counter or evade detection by our system. As adversarial actors gain access to the source code, they could exploit its known strengths and weaknesses to develop countermeasures that diminish its effectiveness. However, despite these risks, we believe that open-sourcing remains the right path forward. Open-sourcing encourages transparency, collaboration, and rapid innovation, enabling the broader community to contribute improvements, detect vulnerabilities, and build on the system.

Moreover, by engaging the community, we can foster the development of more resilient and adaptive models that evolve in response to emerging adversarial techniques, thus maintaining \frameworkname{}’s effectiveness in the long term. The collective strength of a diverse, open-source community can outweigh the potential threats posed by adversarial exploitation.

Future work will need to address these limitations to enhance the practical utility, robustness, and long-term resilience of \frameworkname{}.

\subsection{Sample Image-Caption Pair in the News Domain}\label{app1}
\begin{figure}[H]
    \centering
    \includegraphics[width=0.5\linewidth]{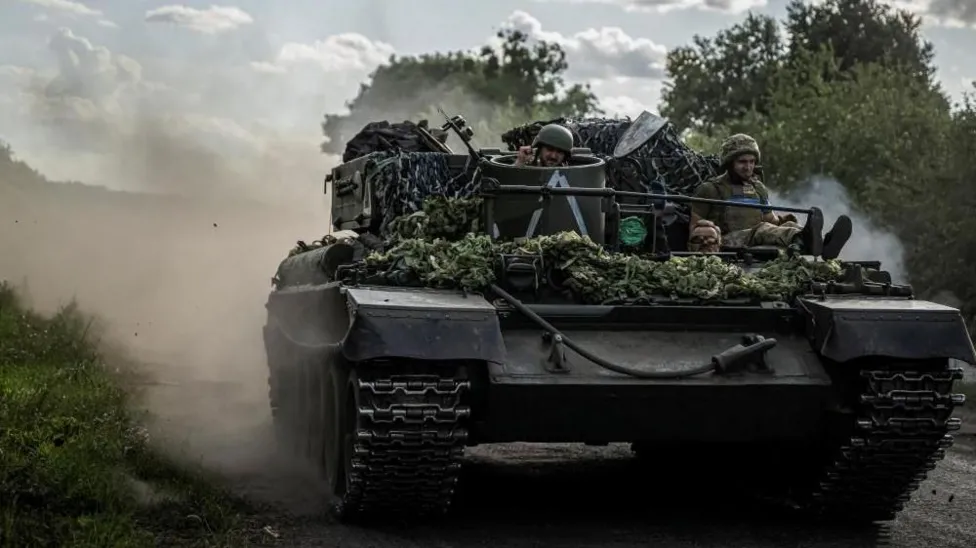}
    \vspace{0.3cm}
    \caption{Russian President Vladimir Putin has called Ukraine's move into Kursk a ``major provocation''. 
    Image and caption taken from the BBC article here (Accessed at 17:43 on Aug 11, 2024): \url{https://www.bbc.co.uk/news/articles/cze5pkg5jwlo}}
    \label{fig:app_1_fig}
\end{figure}

\subsection{Prompts for \frameworkname{}}\label{app:prompts}
\begin{figure}[H]
    \centering
    \framebox{\parbox{\dimexpr\linewidth-2\fboxsep-2\fboxrule}{\texttt{You are a misinformation detection expert in the news domain. You will look at image-caption pairs and decide if the given image is rightly used in the given news context. To further assist you, a summary of news articles related to the image will be provided. Based on this, you need to decide if the caption belongs to the image or if it is being used to spread false information to mislead people. Note that the image is real. It has not been digitally altered. Carefully examine the image for any known entities, people, watermarks, dates, landmarks, flags, text, logos and other details which could give you important information to better explain your answer. Remember in news articles images and captions are often related contextually and the caption need not exactly describe the image. The goal is to consider the contextual relationship between the image and caption based on the news articles and correctly identify if the image caption pair is misinformation or not and to explain your answer in detail. Think step by step and plan a detailed explanation for your answer.}}}
    \vspace{0.1cm}
    \caption{Model initialization prompt.}
    \label{fig:prompt0}
\end{figure}

\begin{figure}[H]
    \centering
    \framebox{\parbox{\dimexpr\linewidth-2\fboxsep-2\fboxrule}{\texttt{This is a summary of news articles related to the image: \{\}\\
                    Based on this, you need to decide if the caption given below belongs to the image or if it is being used to spread false information to mislead people.\\
                    CAPTION: \{\}\\
                    Note that the image is real. It has not been digitally altered. \\
                    Carefully examine the image for any known entities, people, watermarks, dates, landmarks, flags, text, logos and other details which could give you important information to better explain your answer.\\
                    The goal is to correctly identify if this image caption pair is misinformation or not and to explain your answer in detail.\\
                    At the end give a definite YES or NO answer to this question: \\IS THIS MISINFORMATION?}}}
    \vspace{0.1cm}
    \caption{Initial prompt for independent opinion formation and response generation with GPT-4o.}
    \label{fig:prompt1}
\end{figure}

\begin{figure}[H] 
    \centering
    \framebox{\parbox{\dimexpr\linewidth-2\fboxsep-2\fboxrule}{\texttt{This is what I think: \{\}. \\
    Do you agree with me?\\ If you think I am wrong then convince me why you are correct.\\
            Clearly state your reasoning and tell me if I am missing out on some important information or am making some logical error.\\
            Do not describe the image. \\
            At the end give a definite YES or NO answer to this question:\\ IS THIS MISINFORMATION?}}}
   \vspace{0.1cm}
    \caption{Prompt for first round of debate.}
    \label{fig:round_1_prompt}
\end{figure}

\begin{figure}[H] 
    \centering
    \framebox{\parbox{\dimexpr\linewidth-2\fboxsep-2\fboxrule}{\texttt{I see what you mean and this is what I think: \{\}. \\Do you agree with me?\\
                If not then point out the inconsistencies in my argument (e.g. location, time or person related logical confusion) and explain why you are correct. \\
                If you disagree with me then clearly state why and what information I am overlooking.\\
                Find disambiguation in my answer if any and ask questions to resolve them.\\
                I want you to help me improve my argument and explanation. \\
                Don't give up your original opinion without clear reasons, DO NOT simply agree with me without proper reasoning.\\
                At the end give a definite YES or NO answer to this question:\\ IS THIS MISINFORMATION?}}}
     \vspace{0.1cm}                
    \caption{Prompt for subsequent rounds of debate.}
    \label{fig:debate_prompt}
\end{figure}

\subsection{Supplemental Results}\label{app:results}

\subsubsection{Additional Results on VERITE}
\begin{table}[ht]
  \centering
  \begin{tabular}{ll l c c c}
    \toprule
    \textbf{Model} & \textbf{External Retrieval} & \textbf{Data} & \textbf{Accuracy} & \textbf{Precision} & \textbf{Recall}  \\
    \midrule
    4o    & \cmark                & VERITE (all)            & 85.2   & 83.6    & 96.0  \\
    4o    & \cmark                & VERITE (true vs.\ OOC)  & 79.5   & 69.7    & 96.4  \\
    4o    & \cmark                & VERITE (true vs.\ MC)   & 80.7   & 73.6    & 95.6  \\
    4o    & \xmark                 & VERITE (all)            & 84.8   & 85.0    & 93.1  \\
    4o    & \xmark                 & VERITE (true vs.\ OOC)  & 78.6   & 72.1    & 89.5  \\
    4o    & \xmark                 & VERITE (true vs.\ MC)   & 82.6   & 75.6   & 96.4  \\
    \bottomrule
  \end{tabular}
  \vspace{0.1cm}
  \caption{\textbf{External Retrieval Ablation on VERITE.} We compare performance of LLM-Consensus with and without external retrieval with a GPT-4o backbone. We also disaggregate our results by ``miscaptioned`` (MS) and ``out-of-context`` (OOC), two different classes of misinformation included in the VERITE dataset.}
  \label{tab:veriteb-results}
\end{table}

\subsubsection{Additional Baselines for NewsCLIPings}
\begin{table}[h]
\centering
\resizebox{0.4\linewidth}{!}{
\begin{tabular}{l c}
\toprule
\textbf{Model} & \textbf{Accuracy $\uparrow$}\\
\midrule 
SAFE & 50.7 \\
EANN & 58.1 \\
VisualBERT & 54.8\\
CLIP & 62.6\\
InstructBLIP & 48.6 \\
LLaVA & 57.1 \\
GPT-4o & 70.7 \\
DT-Transformer & 77.1 \\
CCN & 84.7 \\
SSDL & 65.6 \\
VINVL & 65.4  \\
Neuro-Sym & 68.2 \\
GPT-4o\textsuperscript{\#} (w retrieval) & 86.0 \\
Sniffer (w finetuning) & 88.4 \\
Sniffer (w/o finetuning) & 84.5 \\
\textbf{\frameworkname{} (ours)} & \textbf{90.8} \\
\bottomrule
\end{tabular}
}
\vspace{0.1cm}
\caption{\textbf{Performance comparison on NewsCLIPpings against various legacy methods: }\frameworkname{} (with GPT-4o) outperforms all related work. Note: the GPT-4o\textsuperscript{\#} setup is an ablation in which there is a single agent using our external retrieval module (no debate).}
\label{tab:baseline_comparison}
\end{table}

\subsection{Qualitative Analysis}\label{app:qualitative}

\begin{figure}[H] 
    \centering
    \framebox{\parbox{\dimexpr\linewidth-2\fboxsep-2\fboxrule}{%
{\ttfamily
Image Description: A banner that says ``STOP VICTIMISATION OF UNION MEMBERS END THE BLACKLIST'' in front of a city landscape.\\

Caption: Flashmob protest at Crossrail site in Central London by Unite union members in protest against blacklisting and sackings of union officials and members.\\

Label: This is not misinformation.\\

Framework prediction: This is not misinformation.\\

Explanation (Model Output): Here is a precise breakdown:\\

1. Image Analysis:\\
- The banner in the image clearly states ``STOP VICTIMISATION OF UNION MEMBERS END THE BLACKLIST.''\\
- This message aligns with union-related grievances focused on stopping blacklisting and mistreatment of union members.\\

2. Caption Analysis:\\
- The caption explicitly mentions a flashmob protest by Unite union members at a Crossrail site in Central London.\\
- The topics of blacklisting and the sacking of union officials and members are mentioned in the caption, aligning with the grievances captured in the image.\\

3. News Summary Correlation:\\
- The true news articles discuss blacklisting practices within the construction industry, particularly concerning the Crossrail project and actions taken by the Consulting Association.\\
- Gail Cartmail from Unite the Union has provided evidence of blacklisting affecting union members working on the Crossrail project, which fits the context depicted in the banner.\\
- The articles also highlight broader issues of blacklisting, further supporting the theme reflected in the image and caption.\\

4. Contextual Match:\\
- The image and the caption are aligned in context. The message on the banner directly supports the protest described in the caption.\\
- The news article summary corroborates the issues of blacklisting faced by union members, as illustrated by the protest.\\

Your reasoning is logically sound, and no critical information is missing. Both the image and the caption are contextually connected to the theme of union protests against blacklisting, specifically within the Crossrail project setting.\\

Thus, I conclude:\\

Definite Answer: NO, the image-caption pair does not represent misinformation.
}}}
     \vspace{0.1cm}                
    \caption{Qualitative sample of model explanation.}
    \label{fig:sample_response}
\end{figure}

\subsection{Ablations}\label{app:ablations}
\subsubsection{Ablations}

To analyze the importance of each component of the \frameworkname{} framework, we conduct ablation experiments. Specifically, we evaluate the effect of using LLaVA against GPT-4o, the impact of the external retrieval module, and the power of the multi-agent debate framework.

\begin{table}[t]
    \centering
    \resizebox{0.99\linewidth}{!}{
    \begin{tabular}{cccc|ccc}
        \toprule
         \textbf{LLaVA} & \textbf{GPT-4o} & \textbf{Retrieval} & \textbf{Debate} & \textbf{Accuracy} & \textbf{Precision} & \textbf{Recall} \\ 
        \midrule
         \cmark & \xmark & \xmark & \cmark & $77.1$ & $68.4$ & $89.3$ \\ 
         \cmark & \xmark & \cmark & \cmark & $86.2$ & $82.6$ & $90.6$ \\ 
         \xmark & \cmark & \cmark & \xmark & $86.0$ & $80.2$ & $95.6$ \\ 
         \xmark & \cmark & \xmark & \cmark & $90.2$ & $\mathbf{90.3}$ & $90.1$ \\
         \xmark & \cmark & \cmark & \cmark & $\mathbf{90.8}$ & $85.5$ & $\mathbf{99.0}$ \\
        \bottomrule
    \end{tabular}}
    \vspace{0.1cm}
    \caption{\textbf{Ablation:} Quantitative evaluation of each component of \frameworkname{} on NewsCLIPpings classification performance.}
\end{table}

We observe that the combination of GPT-4o, the external retrieval module, and the multi-agent debate framework yields the highest performance across all metrics, with 90.8\% accuracy, 85.5\% precision, and 99.0\% recall, demonstrating the value of combining these components. The inclusion of debate alone significantly boosts accuracy from the GPT-4o baseline of 70.7\% (as seen in Table~\ref{tab:baseline_comparison}) to 90.2\%, underscoring its role in enabling contextual reasoning and refining predictions. Adding external retrieval to the GPT-4o with debate system primarily shifts the balance between precision and recall, where precision moves from 90.3\% to 85.5\% and recall from 90.1\% to 99.0\%. Meanwhile, retrieval contributes more substantially to LLaVA’s performance gains, likely due to GPT-4o’s broader world knowledge. Without external retrieval or the debate framework, the performance drops, emphasizing the critical role of these components in achieving state-of-the-art results. We show additional results on VERITE when ablating the retrieval module in Appendix~\ref{app:results}.

\subsection{User Study}\label{app:user_study}
We conduct a user study to assess the effectiveness of our model in detecting and explaining misinformation. Through this study, we aim to assess the persuasiveness of our system.

\subsubsection{Study Design}
The user study was designed to evaluate the effectiveness of our system in detecting and explaining misinformation. While it is easy to quantify model performance in terms of misinformation detection, there are no effective metrics to assess the quality of the explanations generated by the model. Therefore, in order to perform a thorough analysis of the system performance, a user study is essential.

A total of 30 participants volunteered to participate in this study. Participation was completely voluntary and no personal information was used for the purpose of analysis in this study. For a deeper analysis we further grouped the participants based on their profession into three groups, namely: Journalists, AI Academics and Others. The `others' category included anyone who did not belong to the first two groups. The study was conducted through a Microsoft Form. Participants were shown 10 image-text pairs and were asked to decide if the image and caption when considered together was misinformation or not. They were also asked to provide a confidence rating for their answer on a scale of 0-10, with 10 being the highest confidence level. For each image-text pair, after the participants provided their initial answers, they were shown AI insights about the same image-text pair. These AI insights were the final outputs from \frameworkname{}. Participants were then asked to reconsider their answer and again decide if the image-text pair was misinformation or not, in light of the new information from the AI agent. Participants were also required to re-evaluate their confidence score in this new answer. While it is not entirely avoidable, we did ask participants to keep aside their personal opinions of AI and consider all AI insights objectively. Participants were not allowed to access the Internet. This was done to ensure an unbiased estimate of average human performance. 

The image-text pairs to include in the study were taken from the NewsCLIPpings \citep{luo2021newsclippings} dataset. AI insights were taken from our best-performing setup involving the GPT-4o model. Of the 10 image-text pairs presented to the participants in the study, there were 5 instances of misinformation and 5 instances of true information. Further, all model insights were true except two of them. Therefore the model accuracy for the task was 80\% and we use this as the baseline accuracy to compare human performance against.


We analyse two special cases, where \frameworkname{} argues for the wrong answer. We include these results in order to observe how persuasive our system can be even when it is wrong. We note in the instance where the image-text pair was actually misinformation and the model argued that it was not, 6 participants changed their correct responses to those suggested by \frameworkname{}. Although this is only 5\% of the participants, it still gives a significant insight into how persuasive the model can appear even when it is wrong. While the case of false negatives is important, false positives are an even more concerning matter for our problem statement. In the case where \frameworkname{} declared the given image-text pair to be misinformation when it was not, is important to analyse. In this setting 50\% of the total participants changed their answer to the wrong one, therefore believing a piece of true information to be false. In some cases where participants chose the wrong response to begin with, their confidence in the response further increased after considering insights from the system. Finally, 4 participants did not change their answer to the wrong one after considering AI insights but their confidence in their response decreased.

The average time taken to complete the study was 12 minutes and 57 seconds. The average participant was therefore able to go through 10 image-text pairs and decide if they were misinformation or not in under 13 minutes. The same task without AI insights would require extensive analysis and we project it would take between 30-45 minutes to decide if 10 image-text pairs were misinformation.

\subsection{Screenshots}

We include representative screenshots of the user study. 

\begin{figure}[ht]
    \centering
    \begin{subfigure}{0.49\linewidth}
        \includegraphics[width=\linewidth]{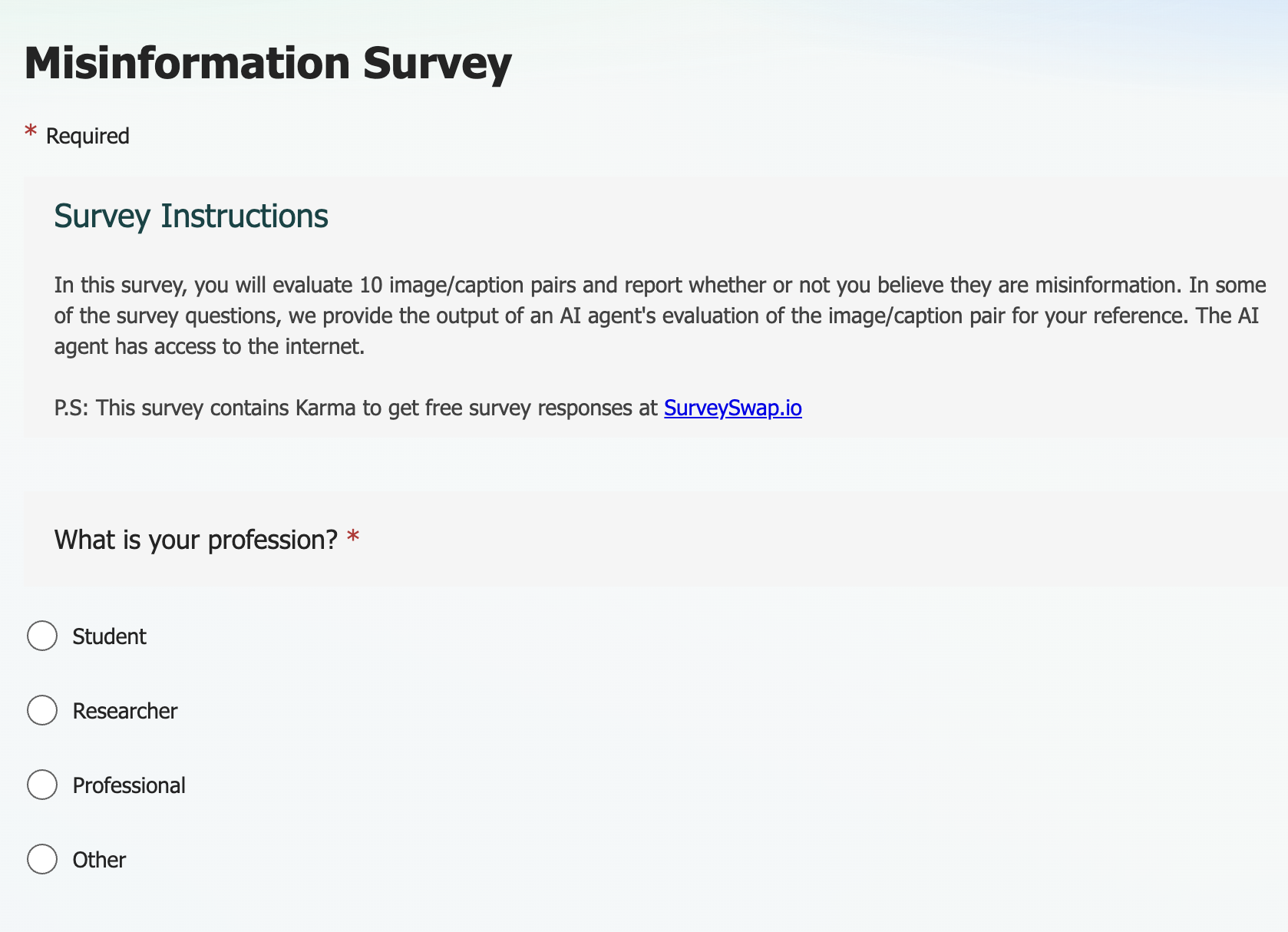}
        \caption{Instructions screen}
        \label{fig:instructions}
    \end{subfigure}
    \hfill
    \begin{subfigure}{0.49\linewidth}
        \includegraphics[width=\linewidth]{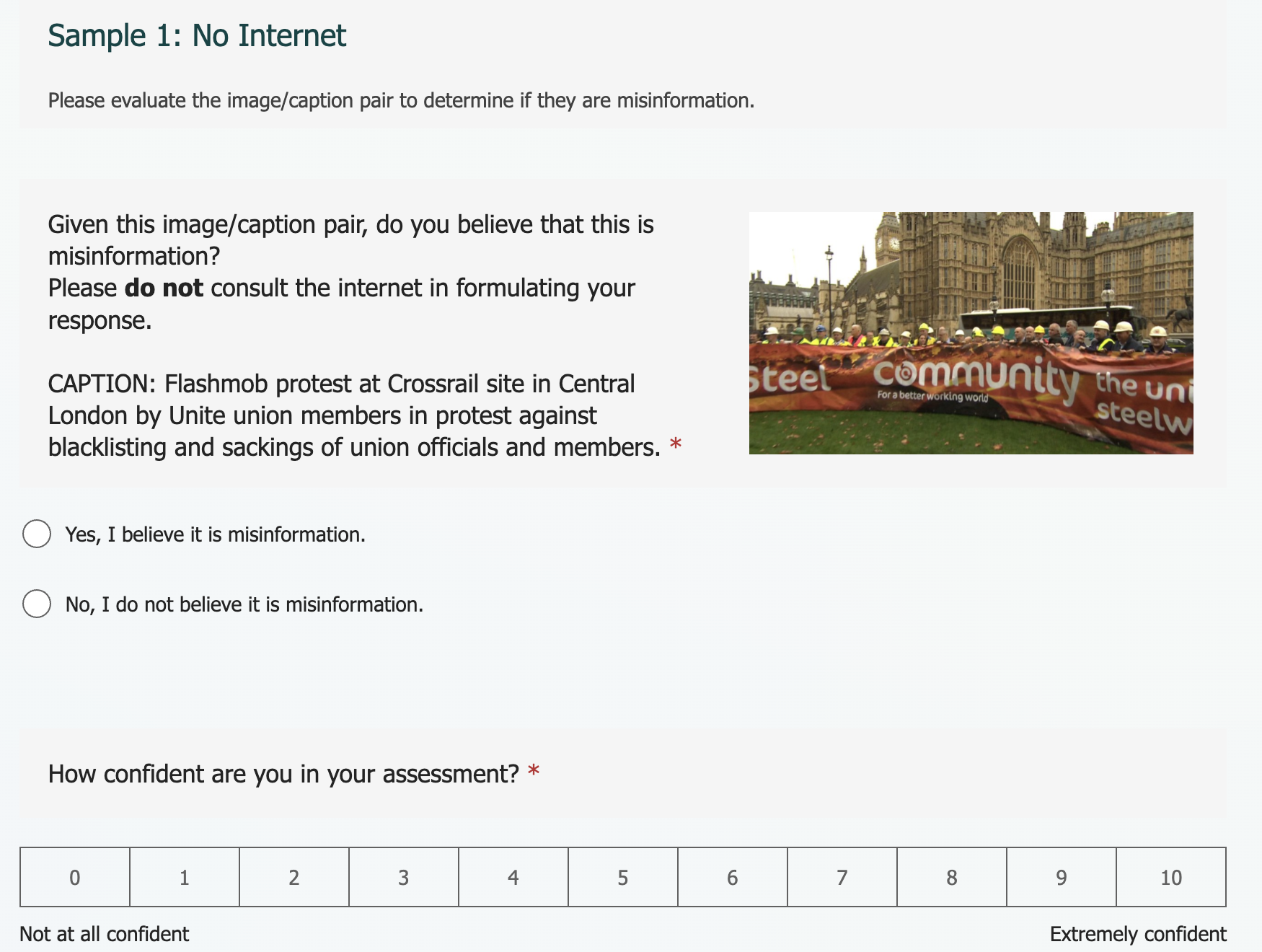}
        \caption{No internet baseline}
        \label{fig:no_internet}
    \end{subfigure}

    \vspace{1em}

    \begin{subfigure}{0.49\linewidth}
        \includegraphics[width=\linewidth]{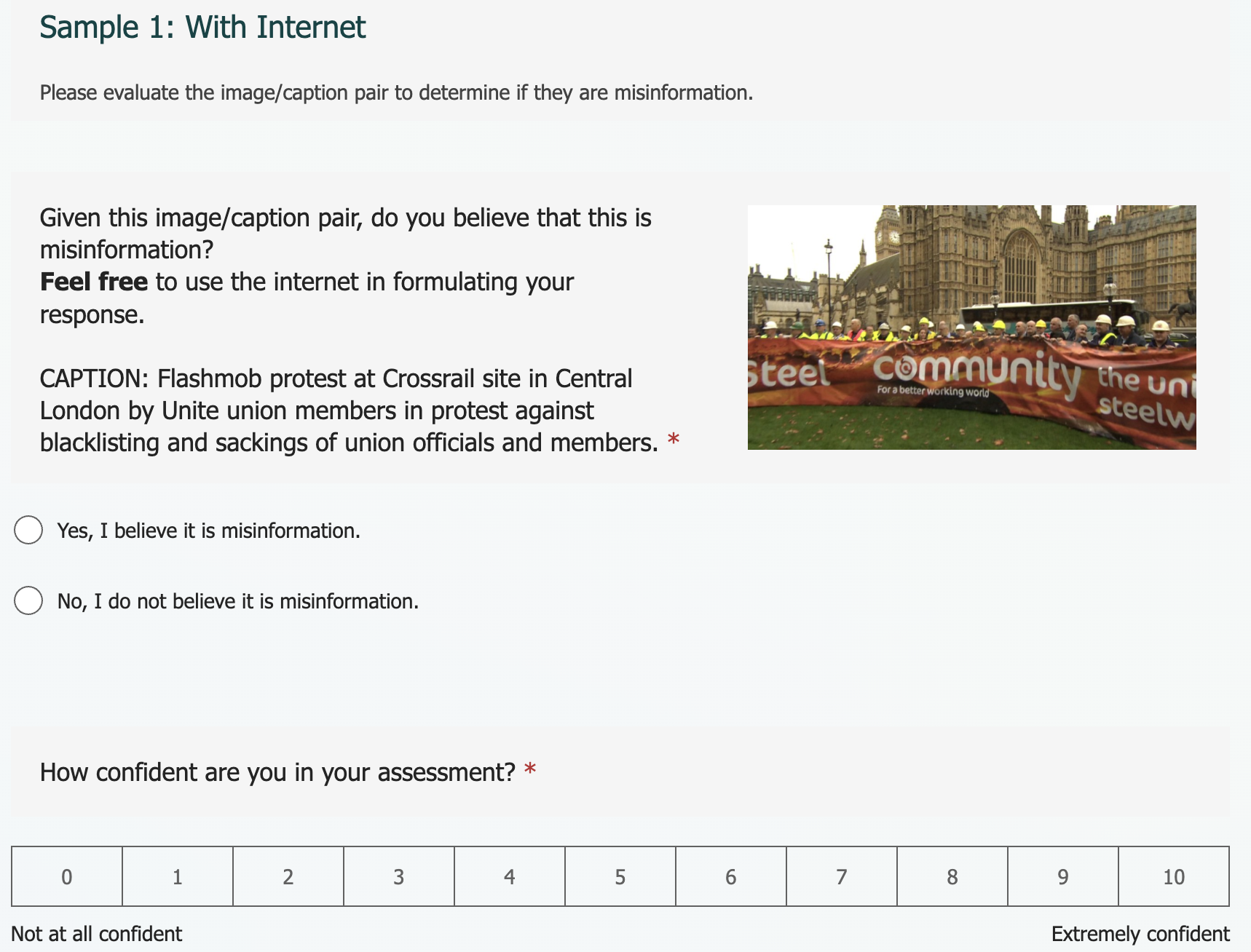}
        \caption{With internet baseline}
        \label{fig:with_internet}
    \end{subfigure}
    \hfill
    \begin{subfigure}{0.49\linewidth}
        \includegraphics[width=\linewidth]{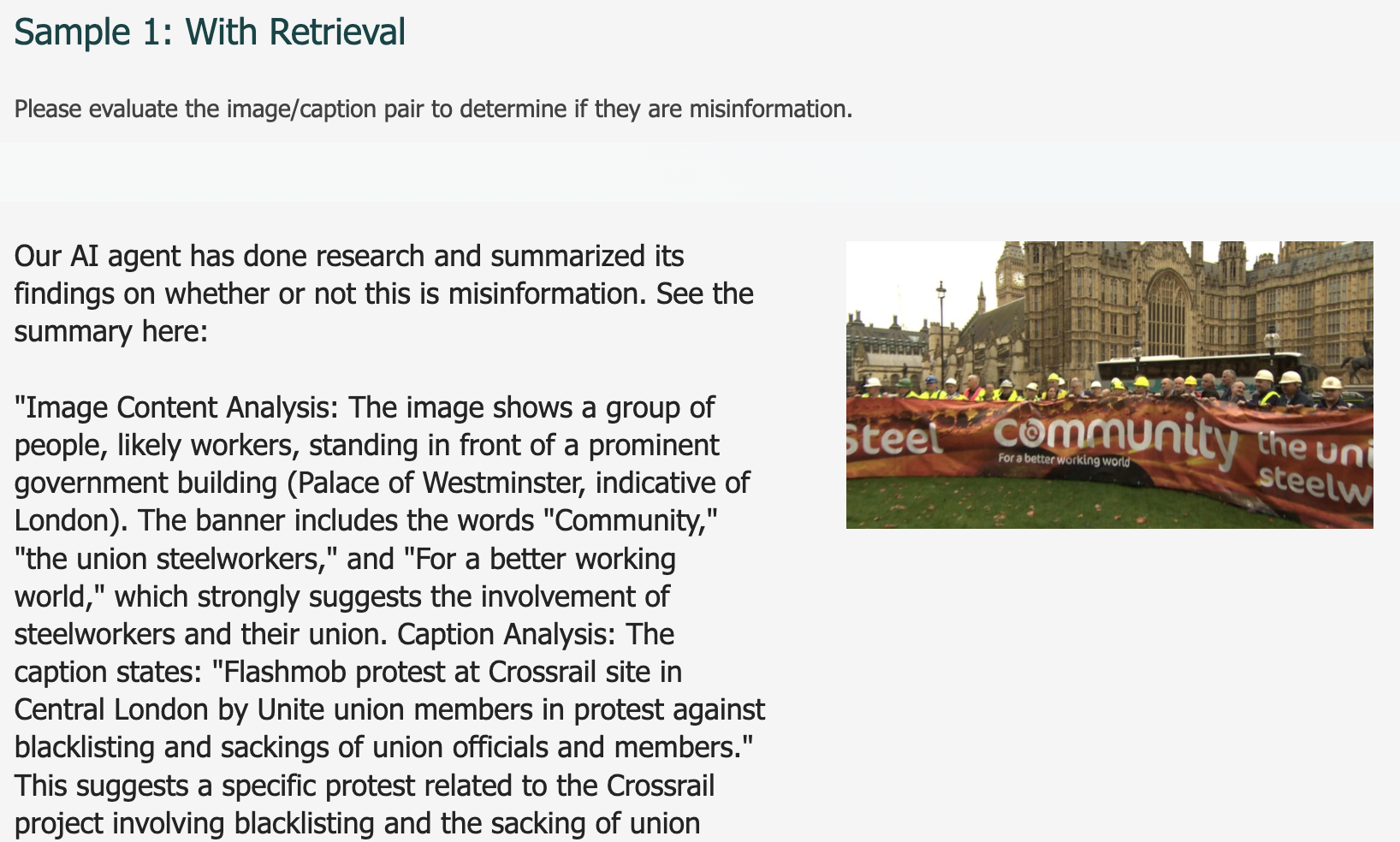} 
        \caption{With retrieval-augmented summary}
        \label{fig:with_retrieval}
    \end{subfigure}

    \caption{Screenshots of the user survey. The questions asked after AI summary is presented are the same as those following the other questions.}
    \label{fig:survey_screenshots}
\end{figure}

\subsection{Multi-modal debates for harmful meme detection}
While this work relates to a different problem than OOC misinformation detection in the news domain, we still find the approach taken by the authors a relevant related work and therefore include it here.
\citet{lin2024explainableharmfulmemedetection} use LMMs debating against each other to generate explanations for contradictory arguments regarding whether a given meme is harmful. These explanations are then used to train a small language model as a judge to determine whether the image and text that make up the meme are actually harmful. 
This work does not allow agents to have flexibility of opinion. There are always two agents, and each one is provided a stance to defend. Moreover, a judge decides the final outcome of the debate and needs to be trained on data from the debate. This method also does not benefit from external retrieval, and therefore, the debating agents are not aware of the crucial external context related to the input. Finally, this work is related to harmful \textit{meme} detection and does not concern the problem of misinformation detection in the news domain, which likely requires more intricate contextual analysis, including of external context.

\subsection{Additional Experiments}\label{app:additional}

\paragraph{Debate with Disambiguation:} Building on the actor-skeptic method, we allow all agents to act as both actors and skeptics. Models generate their own responses and disambiguation queries to refine or challenge other agents' outputs. These queries are used to retrieve additional information from the Internet, further improving model responses. The Debate with Disambiguation strategy achieves accuracy of 77.8, precision of 74.7, and recall of 82.6 when tested with a LLaVa backbone.

\end{document}